# Distilling BERT for low complexity network training

**Bansidhar P.M.**
**bansidhar.mangalwedhekar@nordicsemi.no**

## Abstract

This paper studies the efficiency of transferring BERT learnings to low complexity models like BiLSTM, BiLSTM with attention and shallow CNNs using sentiment analysis on SST-2 dataset. It also compares the complexity of inference of the BERT model with these lower complexity models and underlines the importance of these techniques in enabling high performance NLP models on edge devices like mobiles, tablets and MCU development boards like Raspberry Pi etc. and enabling exciting new applications.

## 1 Introduction

Transfer learning and distilling knowledge from BERT, a state-of-the-art language representation model, into lower dimensional neural networks (BiLSTMs, lower complexity BERT-type models etc) is a well-researched topic (Raphael Tang et. al., 2019, Manish Munikar et. al, 2019, Zhenzhong Lan,2020 etc) and has proven to give very good results across specific tasks at a much lower model complexity and inference time, compared to the BERT only models. This is proven across multiple datasets in paraphrasing, natural language inference and sentiment classification.

The result and comparisons shown in Table II of Manish Munikar et. al, 2019, shows the accuracy of various models on SST-2 and SST-5 using a simple Softmax classifier above BERT. It includes results for all phrases as well as for just the root (whole review). It demonstrates that this model, despite being a simple architecture, performs better in terms of accuracy than many popular and sophisticated NLP models. Raphael Tang et. al., 2019, have demonstrated that lightweight neural networks can be toubo-charged by distilling from BERT.

This paper is an extension of this work and evaluates the efficiency of adding a self-attention layer on top of BiLSTM and also presents the effects of distillation on shallow CNNs.

The hypothesis here is that we should see an improved performance on these low complexity models with distillation via pre-trained BERT teacher mode. A secondary hypothesis is also that the attention layer added on top of BiLSTM should inherently improve its performance over vanilla BiLSTM and should improve further with distillation.

One difference from the reference paper (Raphael Tang et. al., 2019) here is the use of even lower complexity of the models and absolutely no pre-training of the test models either via pre-trained word embeddings or use of POS etc to augment the input sentences. The comparison is primarily based on binary sentiment analysis across the models.

Based on the results of this study, we see an improvement of ~2.2% in accuracy by adding attention to the BiLSTM network. Also, we see about 1% – 1.5% improvement in accuracy for CNN and BiLSTM models by the distillation process. However, we do not see any improvement in the BiLSTM with attention model with distillation. We attribute this result to the use of very low complexity model parameters and a notion of saturation setting in during distillation attributing to this result. An analysis of number of parameters of BERT (base) versus the low-complexity models reveal that these are an order of 175 times lower than the BERT(base) model.

The rest of the paper is arranged in five sections. Section 1 gives an overview of the related prior work. Section 3 introduces the details of the models used and the distillation mechanism. Section 4 details the data used, experimental setup and evaluation methodology. Section 5 tabulates the results of the various experiments. In Section 6, we analyze the results of the different experiments and in Section 7, we provide the conclusions and future work.

## 2 Related work

Raphael Tang et. al., 2019, demonstrate that rudimentary, lightweight neural networks can still be made competitive without architecture changes, external training data, or additional input features. It proposes to distill knowledge from BERT, a SOTA language representation model, into a single-layer BiLSTM, as well as its Siamese counterpart for sentence-pair tasks. The authors propose a simple yet effective approach that transfers task-specific knowledge from BERT to a shallow neural architecture—in particular, a bidirectional long short-term memory network (BiLSTM). Manish et. al., 2019, used a very simple softmax classifier above BERT to solve the fine-grained sentiment classification task on the Stanford Sentiment Treebank (SST) dataset.

Zhenzhong Lan et. al., proposed ALBERT, a lightweight BERT model. Design of ALBERT involved identifying the dominant driver of NLP

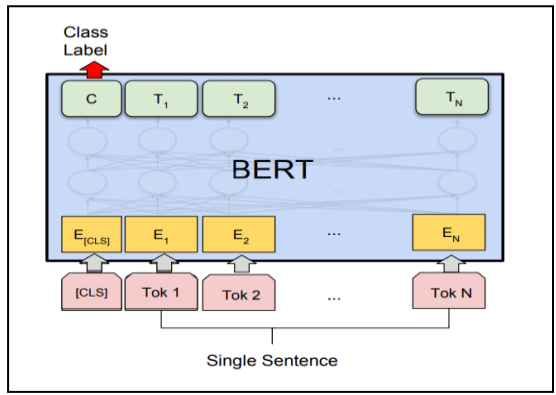

Figure 1: BERT block diagram

performance and allocating the model's performance more efficiently Input-level embeddings (words, sub-tokens, etc.) learn context-independent representations while hidden-layer embeddings refine that into context-dependent representations. While BERT relies on independent layers stacked on each other, ALBERT eliminates this redundancy by applying the same layer on top of each other. Parameter sharing achieves a 90% parameter reduction for the attention-feedforward block (a 70% reduction overall), which, when applied in addition to the factorization of the embedding parameterization, incur a slight performance drop of -0.3 on SQuAD2.0 to 80.0, and a larger drop of -3.9 on RACE score to 64.0. Implementing these two design changes together yields an ALBERT-base model that has only 12M parameters, an 89% parameter reduction compared to the BERT-base model, yet still achieves respectable performance across the benchmarks considered.

Zhang, Y et. al., 2015, conducted a sensitivity analysis of one-layer CNNs to explore the effect of architecture components on model performance. Their aim was to distinguish between important and comparatively inconsequential design decisions for sentence classification. The study is useful for getting the most out of CNNs for sentence classification in real world settings.

In summary, all these works studied optimizations primarily with BERT in-mind to achieve a design target. The present study is an extension of Raphael Tang et. al., 2019, where we examine the improvement in performance by adding attention over the BiLSTM model, evaluating a shallow CNN mode and implementing the distillation process to evaluate its efficiency in training these low-complexity models with the pre-trained BERT teacher model.

## 3 Approach

We consider Bi-directional LSTM (BiLSTM), BiLSTM with Attention and shallow CNN networks for our evaluation. We describe all the models in brief and then the distillation methodology used for transferring the BERT knowledge into these lower models.

### 3.1 BERT

BERT (Jacob Devlin et. al, 2018) is the recent addition to the NLP pre-training algorithms, leveraging large amounts of existing text to pretrain a model's parameters using self-supervision, with no data annotation required. It can be used for extracting high quality language features from test data as well as fine-tune for tasks like sentiment analysis or question answering to produce SOTA results. BERT advances the state of the art for eleven NLP tasks

BERT's model architecture (Figure 1: BERT block diagram) is a multi-layer bidirectional Transformer encoder. Since its goal is to generate a language representation model, it only needs the encoder part. Looking at context around the masked word from both left and right sides at the same time is one of the key differences from a bi-direction RNNs.

The key innovative part of BERT is that it is context based and bidirectionally trained. The pre-training is primarily based on two tasks –

1.      MLM - Instead of predicting the next word in a sequence, BERT makes use of a novel technique called Masked LM - it randomly masks words in the sentence and then it tries to predict them. With masking, the model looks in both directions and it uses the full context of the sentence, both left and right surroundings , in order to predict the masked word. This is called Cloze task in literature. In practice, 15% of token positions are taken in random, of which 80% are masked with [MASK], 10% are replaced with random tokens and 10% are left unchanged – this mitigates mismatch between pre-training and fine-tuning phases.

2.      NSP - Many important downstream tasks such as Question Answering (QA) and Natural Language Inference (NLI) are based on understanding the relationship between two sentences, which is not directly captured by language modeling. In order to train a model that understands sentence relationships, we pre-train for a binarized *Next Sentence Prediction* task. In order to train a model that understands sentence relationships, we pre-train for a binarized next

There are two steps in the training framework: pre-training and fine-tuning. During pre-training, the model is trained on unlabeled data over different pre-training tasks. For finetuning, the BERT model is first initialized with the pre-trained parameters, and all of the parameters are fine-tuned using labeled data from the downstream tasks.

BERT used WordPiece embeddings (Wu et al.,2016) with a 30,000 token vocabulary , the BooksCorpus (800M words) (Zhu et al.,2015) and English Wikipedia (2,500M words) for pre-training of various aspects.

Experiments conducted to explore the effect of model size on the fine-tuning task accuracy, concluded that when the model is fine-tuned directly on the downstream tasks and uses only a very small number of randomly initialized additional parameters, the task specific models can benefit from the larger, more expressive pre-trained representations even when downstream task data is very small. BERT primarily has two model sizes: BERT(base) (L=12, H=768, A=12, Total Parameters=110M) and BERT(large) (L=24, sentence prediction task that can be trivially generated from any monolingual corpus. Specifically, when choosing the sentences A and B for each pretraining example, 50% of the time B is the actual next sentence that follows A (labeled as IsNext), and 50% of the time it is a random sentence from the corpus (labeled as NotNext).

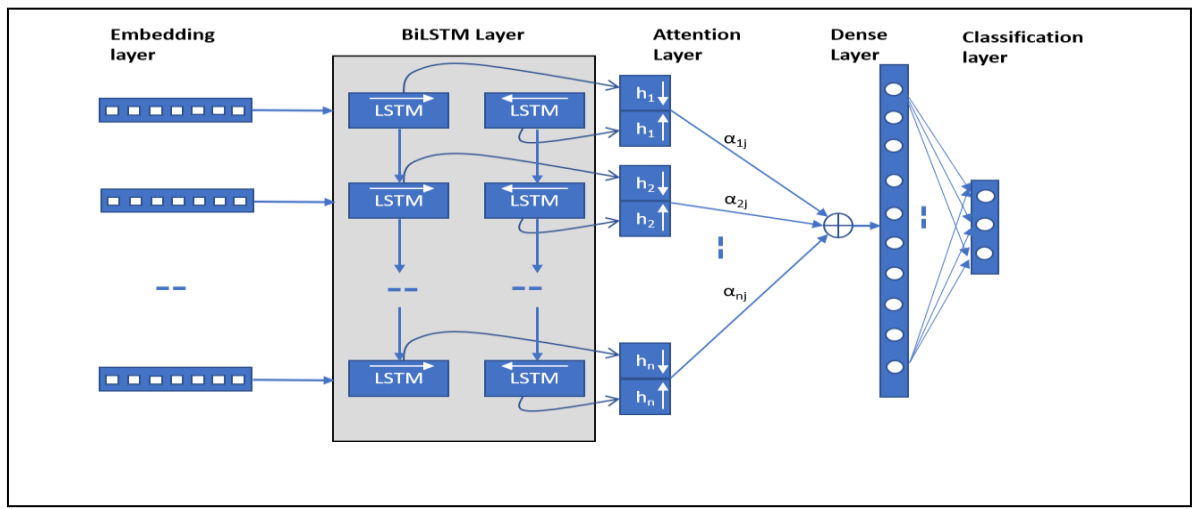

Figure 2: BiLSTM with attention

A distinctive feature of BERT is its unified architecture across different tasks. There is minimal difference between the pre-trained architecture and the final downstream architecture.

H=1024, A=16, Total Parameters=340M). All results show BERT outperforming others to produce STOA results.

## 3.2 BiLSTM

Bidirectional LSTM is an extension of traditional LSTM to train two LSTMs on the input sequence. The second LSTM is a mirror image of the first one, so that we can learn from both past and future input features for a specific time step. We train the Bidirectional LSTM networks using back-propagation through time. After the embedding layer, the sequence of word vectors is fed into the Bidirectional LSTM where the sequence of words in the sentence are passed simultaneously from both directions and a concatenated output and hidden layer representation is available to be used with a fully connected neural network layer to provide the logits to the classification layer to take a binary decision.

to use filters with widths equal to the dimensionality of the word vectors (d). Due to the sequential nature of the words in a sentence, the filters used in text processing are 1d-filters and move only in one direction. We can define multiple such filters to learn different features from the same region of the matrix. We can then do a 1d-max pooling followed by flattening and input to a dense layer that is then inout to a softmax classifier to generate the final classification.

A very shallow/simplistic model based on that is show in Figure 4 : CNN architecture , that is used in our experiment. The input layer is a sentence comprised of concatenated word embeddings. That is followed by a convolutional layer with multiple filters, then a 1d max-pooling layer, and finally a fully connected dense layer with a softmax

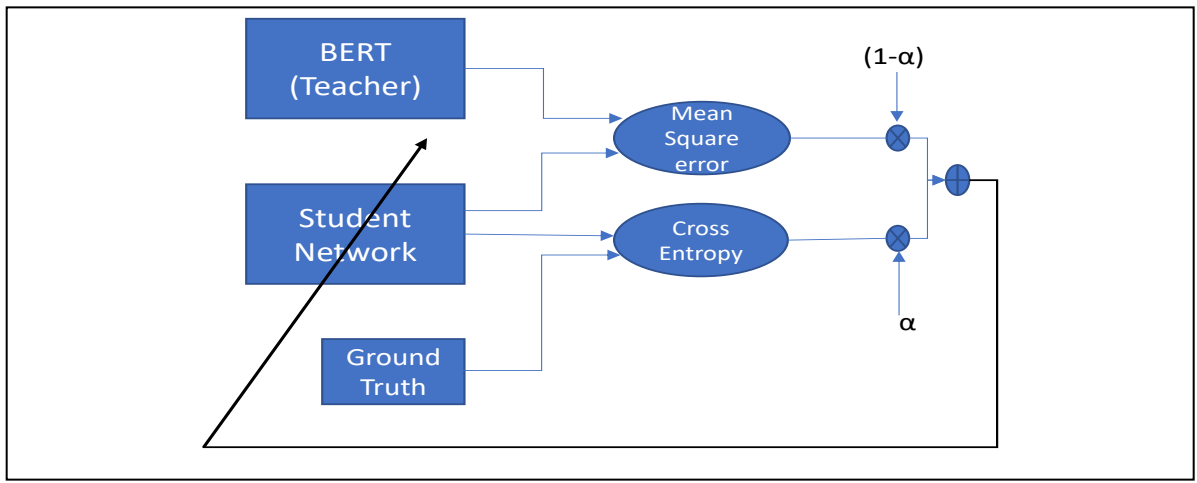

Figure 3 : Distillation block diagram

## 3.3 Shallow CNN

Zhang, Y., & Wallace, B (2015) analyzed CNNs in the context of NLP/text processing. CNNs begin with a tokenized sentence which is converted to a sentence matrix, the rows of which are word vector representations of each token. These might be outputs from trained word2vec (Mikolov et al., 2013) or GloVe (Pennington et al., 2014) models. Let us denote the dimensionality of the word vectors by d. If the length of a given sentence is s, then the dimensionality of the sentence matrix is s × $d^2$. We can effectively treat the sentence matrix as an 'image', and perform convolution on it via linear filters. In text applications there is inherent sequential structure to the data. Because rows represent discrete symbols (words), it is reasonable classifier. A drop-out layer is used to ensure regularization. The complexity of the model is kept simple to ensure very low inferencing time.

## 3.4 Self-Attention

In early NLP, TDIFD/CountVectorizer kind of methods were used to extract keywords and features. However, these are purely bag-of-words kind of statistics and don't preserve any sequence based information vital for text classification. RNNs and deep learning methods take care of the sequence of the texts but in their generic form and give equal weightage to all the words in the sequence. This is where attention mechanism came in and revolutionized NLP.

A typical attention configuration used above the BiLSTM network is shown in Figure 2: BiLSTM with attention.

accumulation of all the h_ij in the sequence following equation 3.

The attention layer is followed by a dense layer with a 2-neuron output layer for binary classification via softmax function giving us the probability of the word belonging to the binary classes.

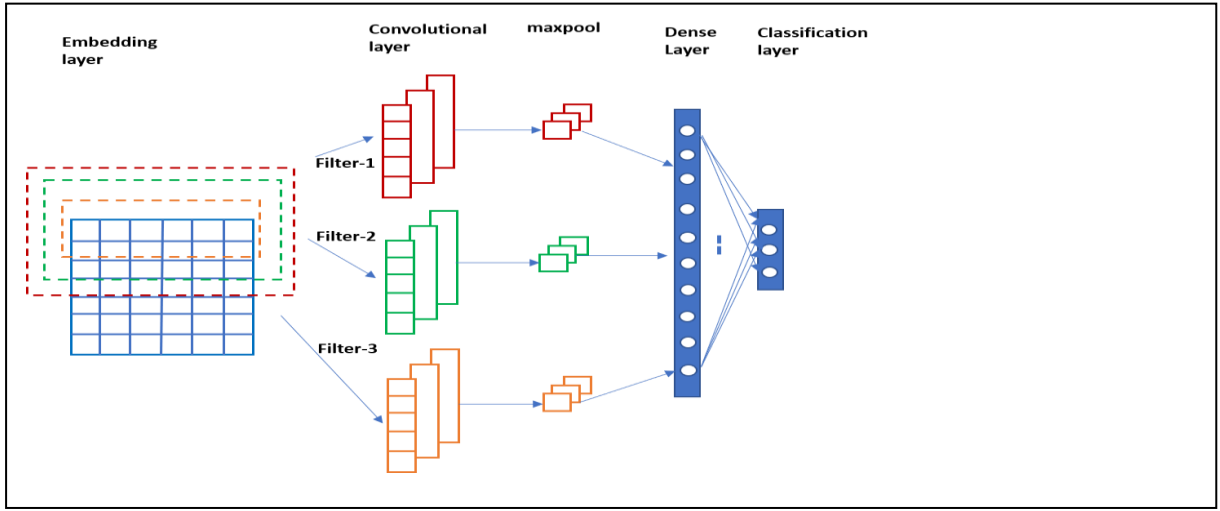

Figure 4 : CNN architecture

### 3.5 Distillation

The process of distillation is depicted in Figure 3 : Distillation block diagram. The teacher mode (BERT in this case) is fine-tuned on the dataset in question. Once that part is complete, the dataset is passed through the student network that is targeted to be trained and the soft logits are then compared to the target soft logits from the teacher and passed through an error function (mean-square error, in this case).

In parallel, a cross-entropy loss is calculated from the output of the student network and the ground truth of the training samples. The cross-entropy error and the mse error are averaged weighted by $\alpha$ and $(1-\alpha)$ and that error is used in back propagation and weights update of the student network.

This also can be thought of as training a student to reproduce the behavior of the teacher as accurately as possible and with fewer parameters

The mean-square error (MSE) – this gives the mean square error of the bert logits, $y_i$ and the estimated logits from the student network $\tilde{y}_i$.

$$MSE = \frac{1}{n}\Sigma_{i=1}^{n}(y_i - \tilde{y}_i) \qquad (4)$$

Binary cross entropy or log-loss function –

The input to the attention layer is the concatenated output (from both directions) from the BiLSTM model.

Let $i$ represent the i$^{th}$ input sequence and $j$ represent the j$^{th}$ word in the i$^{th}$ sequence.

$$u_{ij} = tanh(W h_{ij} + b) \qquad (1)$$

$$\alpha_{ij} = \left(\frac{e^{u_{ij}^T u_w}}{\Sigma_{k=1}^{N} e^{u_{kj}^T u_w}}\right) \qquad (2)$$

$$s_j = \Sigma_k \alpha_{kj} h_{kj} \qquad (3)$$

W is the weight matrix of the linear layer with b being the bias, $\alpha_{ij}$ is the attention weight for the j$^{th}$ word. We now measure the self-attention of the word s_j through the similarity between u_ij and the word level context vector u_w which is randomly initialized.With this, we get a normalized attention vector $\alpha_{ij}$ as the output of a softmax function. $\alpha_{ij}$ is the weight of the j$^{th}$ word in the ith input word sequence. The sequence vector s_j is a weighted

$$H_p(q) == -\frac{1}{N}\Sigma_{i=1}^{N} y_i . \log(p(y_i)) + (1 - y_i) . \log(1 - p(y_i))$$

where y is the ground truth label and p(y) is the predicted probability of the class being 1 for all N points – in this case, the logits from the student network.

# 4   Experiments

## 4.1   Data

We use Stanford Sentiment Treebank (SST, Socher et al., 2013) as the primary dataset. This is a widely-used resource for evaluating supervised NLU models, and one that provides rich linguistic representations. The train/dev/test SST distribution contains files that are lists of trees where the part-of-speech tags have been replaced with sentiment scores 0...4 as below:

- 0 and 1 are negative labels.
- 2 is a neutral label.
- 3 and 4 are positive labels.

This work specifically focuses on **SST-2**, binary set for sentiment classification analysis. Details in Table 1: SST-2 dataset.

| Dataset | #Train | #Dev | #Test | #Classes |
|---------|--------|------|-------|----------|
| SST-2   | 67,350 | 873  | 1822  | 2        |

Table 1: SST-2 dataset

## 4.2   Metrics

### Accuracy -

The SST-2 dataset has both the binary classes uniformly distributed and hence we use 'accuracy' as our evaluation and comparison metric. Accuracy is defined as –

$$Accuracy = (\frac{Number\ of\ correct\ predictions}{Total\ number\ of\ predictions}) \quad (5)$$

### Number of trainable parameters (complexity) -

Comparison of the number of trainable parameters is another metric that we use to compare the models. This has a direct consequence on the inference time and memory requirements on the inferencing device. This comparison is made with respect to the BERT-base model parameters and in terms of a ratio so that one can easily comprehend the comparative size of the reference models. No explicit inferencing time measurements are done.

## 4.3   Evaluation method

Sentiment analysis taken in a sentence at a time, preprocesses it to make it amenable for input to our models, passes that through the models and then makes a classification decision.

Firstly, we run the following baseline models and get an accuracy performance metric for each of them –

1. BERT
2. BiLSTM
3. BiLSTM with Attention
4. CNN

Once we have this, we run each of the models under evaluation (BiLSTM, BiLSTM+Attention, CNN) through distillation process where the trained BERT (base) model will act at the teacher model and each of the evaluation models take the place of the student model.

We then compare the performance of the baseline models before and after distillation process and evaluate the potential performance improvement for each case.

It has been a deliberate design decision to work with very low complexity for the student models that translate to lower inferencing complexity. The details of the parameters used for simulations are as below. The maximum sequence/sentence length is kept at 128 words for all the models.

BiLSTM :

        Embedding dimensions  = 64
        Hidden dimensions     = 64
        Dense Layer           = 1

BiLSTM with Attention:

        Embedding dimensions  = 64
        Hidden dimensions     = 64
        Attention Layer       = 1
        Dense Layer           = 1

CNN :

        Embedding dimensions  = 64
        Hidden dimensions     = 64
        Number of kernels     = 3
        Filters per kernel    = 1
        Dense Layer           = 1

Further note that no pre-defined word embedding (GloVe, Word2Vec, FastText) is used and it is left to the network training to learn these as part of the training process.

Adam optimizer combined with with stepLR and gamma = 0.9 was used throughout this study for baseline and distillation models.

For the baseline models, the "cross entropy loss" function was used and for distillation, a weighted combination of "cross-entropy loss" and "mean-square-error" functionality was used. The mini-batch size of 32 was used throughput the simulations.

The baseline code & formatted SST-2 data used for the simulations is from https://github.com/pvgladkov/knowledge-distillation. Attention, CNN and other additions were coded on top of this reference and code snippets from Stanford cs224u course-work was merged and reused.

Running Simulations –

The codebase for this study is setup such that we first run the standalone BERT test to get the baseline performance metrics. Baseline performance for the other models (BiLSTM, BiLSTM+Attention and CNN) are run via runtime model selection option. Once we have the baseline performance, we run a "distil_bert" with the SST-2 dataset and store the BERT logits to be used as a reference – this is done via the python pickle dump and read functionality. Once we have stored the BERT logits in a pickle file, we simple use this as a reference in the distillation process and use the modified loss functions that take this and the output from the student network to evaluate loss and then feed this back for back-propogation.

Huggingface library is used for BERT training and the pre-trained "bert-base-uncased" is used along with the Bert tokenizer in the simulations. The first time BERT testing is run, it downloads a few GB of pre-trained model data and on subsequent runs, picks this up from the cache.

## 5   Results

Table ABC "Table 2: Accuracy results" shows the

| Model | Baseline | Distillation |
|-------|----------|--------------|
| BERT base | 93.46 | - |
| BiLSTM-Attn | 85.21 | 85.37 |
| BiLSTM | 83.03 | 84.4 |
| CNN | 79.47 | 80.39 |

Table 2: Accuracy results

% accuracy results of the experiments. All simulations are run for 20 epocs and the best model chosen to evaluate on the test dataset.

The table "Table 3: Parameters comparison" gives a comparison of the number of training parameters

| Model | #Params | Inference |
|-------|---------|-----------|
| BERT (base) | 110M | x1 |
| BiLSTM+Attn | 0.75M | x160 |
| CNN | 0.63M | x175 |

Table 3: Parameters comparison

utilized by each model.

We note that our proposed models are orders-of-magnitude lower than the BERT base and hence could easily be inferenced on edge devices. The actual inference time analysis is out of scope of the present study and left as an exercise to a keen researcher.

## 6   Analysis

Reproducing the BERT base results on the SST-2 dataset is successfully accomplished so we know the teacher network is trained as expected. We then set about getting the optimal performance of the student models for the given set of low-complexity parameters. Note here that even though the achieved accuracy of these models is lower than that cited in other papers, this is expected given our chosen hyperparameters. We set this baseline and then repeat the experiments with distillation.

We see an expected trend of improvement in baseline performance from CNN to BiLSTM and then to BiLSTM with attention. We see about 3.5% increase from CNN to BiLSTM and about 2.2% increase going from BiLSTM to BiLSTM with attention. The improvement due to attention mechanism was an expected outcome of the experiment and achieved.

Comparing the baseline performance with the distillation performance (with $\alpha = 0.5$), we see ~1% increase in performance with CNN, ~1.4% increase in performance for BiLSTM but no improvement in performance BiLSTM with attention. While the former two results are on the expected grounds, the fact that we achieved no improvement with distillation on BiLSTM with attention was unexpected but interesting.

After running several tests with changing parameters, we ensured it was not a code bug and this observation was repeatable for each such configuration. The general conclusion here is, since models are low-complexity and the attention mechanism added to BiLSTM already captures a lot of 'word importance' statistics, further distilling did not help improve performance as the model got 'saturated'.

A further detailed study is required in this aspect by going to higher dimensions and hyperparameters while also utilizing more elaborate pre-processing of the input sentences.

## 7    Conclusion

This study implemented attention over BiLSTM network and saw ~2.2% increase in performance due to this. There was a good improvement in performance due to distillation process for all models other than BiLSTM with attention. We observe that while the performance of these low-complexity models was a bit further away from what BERT-base achieved, the primary criteria of the model evaluations was low-complexity and hence the performance of these models is considered good and amenable for fast inference on edge devices. One of the observations after studying through the results is the very obvious revelation that for tasks such as sentiment classification, gigantic models like BERT seems to be an overkill and incorporate a huge volume of redundancy which is basically why comparatively low complexity models come close to performance offered by BERT. We believe there is a lot more to distillation than has been evident in this study and as one dives deeper into the realms of the BERT layers, one might come up with much more optimized architectures to deliver the same punch as BERT at a lot lower cost.

Due to paucity of time, no further enhancements or experiments could be conducted. For a researcher with keep interest in this domain, this work can be extended with higher parameters with the aim of improving accuracy closer to the reference BERT. Also, multiple datasets could be used for comparison. One could also run some real-time inference time & memory requirements on some typical edge devices.

## 8    Acknowledgments

I wish to thank Prof Chris Potts for the very enjoyable topics and format of Stanford's CS224u course, his proactive and expert guidance during this work.